\documentclass[times, twoside]{zHenriquesLab-StyleBioRxiv}

\usepackage{blindtext}  
\usepackage{makecell}   
\usepackage{multirow}   
\usepackage{booktabs}   
\usepackage{cleveref}   
\usepackage{siunitx}    

\crefname{figure}{Fig.}{Figs.}
\Crefname{figure}{Figure}{Figures}
\crefname{table}{Table}{Tables}
\Crefname{table}{Table}{Tables}
\crefname{section}{Sec.}{Secs.}
\Crefname{section}{Section}{Sections}
\crefname{equation}{Eq.}{Eqs.}
\Crefname{equation}{Equation.}{Equations}

\newcommand{\bv}[1]{\mathbf{#1}} 

\bibpunct{(}{)}{;}{a}{,}{,}
\setcitestyle{authoryear,open={(},close={)}}
\renewcommand{\thesubsection}{\thesection.\Alph{subsection}}
\DeclareCaptionFormat{smallformat}{\normalfont\sffamily\fontsize{9}{11}\selectfont#1#2#3}
\renewcommand\Affilfont{\color{color0}\normalfont\sffamily\fontsize{9}{10}\selectfont}

\hypersetup{allcolors=black}
\def\abstract{%
  \setlength{\parindent}{0pt}
  \ifbfabstract\small\bfseries\else\footnotesize\fi}

\makeatletter
\usepackage{fancyhdr}
\newcommand*\BioFooterLeft{%
  \footerfont
  \@leadauthor\ifnum\value{authors}>1\textit{\,et al.}\fi}
\newcommand*\BioFooterRight{%
  \footerfont arXiv\hspace{7pt}|\hspace{7pt}\thepage}

\fancypagestyle{plain}{\fancyhf{}\fancyfoot[LE,LO]{\BioFooterLeft}%
                              \fancyfoot[RE,RO]{\BioFooterRight}}
\makeatother
\makeatletter
\fancypagestyle{plain}{%
  \fancyhf{}                              
  \fancyfoot[RO]{\aBioRXiv}               
  
  }
\makeatother

\leadauthor{Amini} 

\begin{document}
\title{Post-Hurricane Debris Segmentation Using Fine-Tuned Foundational Vision Models}
\shorttitle{}


\author[1,*]{Kooshan Amini}
\author[2,*]{Yuhao Liu}
\author[1,3,\Letter]{Jamie Ellen Padgett}
\author[2,3]{Guha Balakrishnan}
\author[2,3]{Ashok Veeraraghavan}

\affil[1]{~Department of Civil and Environmental Engineering, Rice University}
\affil[2]{~Department of Electrical and Computer Engineering, Rice University}
\affil[3]{~Ken Kennedy Institute, Rice University}
\affil[ ]{{\phantom{\textsuperscript{1}}}\texttt{\{kooshan.amini,yuhao.liu,jamie.padgett,guha,vashok\}@rice.edu}}

\maketitle


\begin{abstract}
\noindent \textit{Abstract} – Timely and accurate detection of hurricane debris is critical for effective disaster response and community resilience. While post-disaster aerial imagery is readily available, robust debris segmentation solutions applicable across multiple disaster regions remain limited. Developing a generalized solution is challenging due to varying environmental and imaging conditions that alter debris' visual signatures across different regions, further compounded by the scarcity of training data. This study addresses these challenges by fine-tuning pre-trained foundational vision models, achieving robust performance with a relatively small, high-quality dataset. Specifically, this work introduces an open-source dataset comprising approximately 1,200 manually annotated aerial RGB images from Hurricanes Ian, Ida, and Ike. To mitigate human biases and enhance data quality, labels from multiple annotators are strategically aggregated and visual prompt engineering is employed. The resulting fine-tuned model, named \emph{fCLIPSeg}, achieves a Dice score of 0.70 on data from Hurricane Ida---a disaster event entirely excluded during training---with virtually no false positives in debris-free areas. This work presents the first event-agnostic debris segmentation model requiring only standard RGB imagery during deployment, making it well-suited for rapid, large-scale post-disaster impact assessments and recovery planning.
\end{abstract}


\begin{corrauthor}
\texttt{jamie.padgett@rice.edu} \\
*~These authors contributed equally.
\end{corrauthor}

\section{Introduction} \label{sec:introduction}

\begin{figure*}
	\centering
	\includegraphics[width=1.0\linewidth]{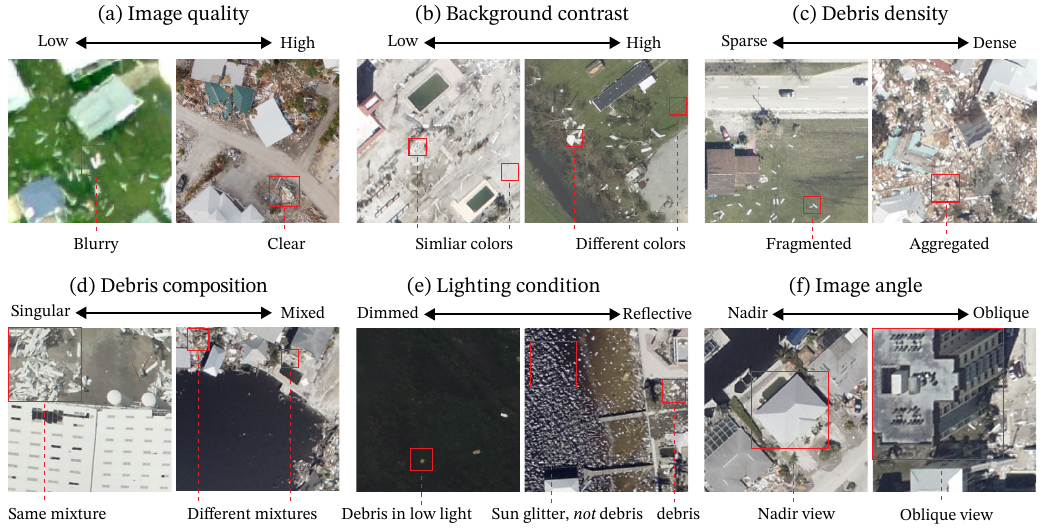}
	\caption{\textbf{Challenges in accurate debris segmentation across diverse conditions.}
    This figure highlights six imaging and environmental factors that alter debris visual signatures. Debris also appears fragmented in aerial imagery with limited visual features, further complicating differentiation from visually similar objects. 
    }
	\label{fig_opening}
\end{figure*}

Natural hazards, particularly hurricanes and coastal storms, are among the most costly and disruptive events, inflicting widespread damage to infrastructure and housing while severely challenging community resilience. In the United States alone, tropical cyclones have caused over \$1.5 trillion in cumulative damages since 1980, averaging approximately \$23 billion per event \citep{NOAA2025}. These disasters not only devastate local economies but also displace hundreds of thousands of people for prolonged periods. A critical consequence following a hurricane is the massive accumulation of debris. For instance, a single major hurricane, such as Hurricane Katrina in 2005, generated over 100 million cubic yards of debris \citep{GAO2008}, placing immense strain on waste management resources. Debris removal often accounts for approximately 36\% of total hurricane-related costs \citep{OIG18-85}. Furthermore, accumulated debris can obstruct key transportation routes and impede emergency response, thereby exacerbating recovery expenses \citep{amini_probabilistic_2023, Zhaojing2018}. Rapid detection and quantification of post-storm debris are thus essential for prioritizing clearance operations, optimizing resource allocation, and ultimately mitigating secondary losses \citep{nickdoost_integrated_2022}.

High-resolution aerial imagery has emerged as a crucial resource for expediting post-disaster assessments \citep{Cheng2021}. Agencies such as the National Oceanic and Atmospheric Administration (NOAA) now routinely deploy aircraft or satellites to capture post-hurricane imagery within days of landfall, offering near real-time overviews of damage and debris distribution \citep{NOAA2018}. This readily available data is invaluable for determining the extent of storm damage and guiding emergency response. However, manually analyzing thousands of aerial images is both time-consuming and labor-intensive, delaying critical decision-making. Consequently, advanced image-processing techniques are required to automate debris detection. Recent studies have leveraged AI-assisted workflows to estimate debris volumes from drone imagery and other remote sensing sources \citep{cheng2024framework}, demonstrating the potential of such methods for rapid, large-scale impact assessments. These developments underscore the urgent need for robust debris segmentation models capable of operating efficiently on aerial imagery with different resolutions and quality to support effective hurricane recovery planning.


Accurate detection of debris across a variety of conditions is challenging, as illustrated in \cref{fig_opening}. First, the visual characteristics of debris are influenced by various factors, including image quality, background contrast, debris density, debris composition, lighting conditions, and image angle, among others.
These factors can vary significantly across different natural hazard events, making consistent identification challenging. Second, due to its fragmented nature, debris may occupy a tiny portion of the image with limited context. As a result, differentiating debris from visually similar objects (such as water reflections or pavement markings) can be difficult. Given these complexities, a robust, generalizable debris segmentation model likely requires a large number of parameters to capture intricate data patterns. However, such models also demand extensive training data \citep{emam2021statedatacomputervision}. While existing programs \citep{noaa_emergency_response_imagery} have facilitated the collection of substantial quantities of post-disaster RGB images, there is currently a dearth of high-quality labeled datasets that could fully support training a large-scale model from scratch.
In the machine learning (ML) community, the scarcity of labeled datasets has driven a shift toward approaches that make use of pre-trained models, enabling researchers to overcome limitations in data availability.

Modern ML methods increasingly rely on \textit{foundational models} to leverage their scalability, pre-trained knowledge, and adaptability across diverse tasks and domains \citep{wang2023exploiting, zhang2023personalize, osco2023segment, panZeroshotBuildingAttribute2023}. 
Foundational models are typically trained on vast amounts of data (reaching the order of billions), resulting in robust data representations and pattern recognition that generalize well to downstream applications. 
Among all foundational models \citep{kirillov2023segment, radford2021learninga}, including their derived image segmentation models \citep{liu2023grounding, ren2024grounded, zhang2023personalize, osco2023segment}, only CLIPSeg \citep{luddecke2022image} demonstrated possible capacity for debris segmentation. 
The CLIP transformer encoder in CLIPSeg is trained on 400 million image-text pairs, enabling learned text-to-image associations on a wide variety of objects and scenes \citep{radford2021learninga}, including remote sensing applications \citep{helber2019eurosat}. 
CLIPSeg \citep{luddecke2022image} utilizes a transformer-based decoder that enables segmentation of arbitrary text and visual prompts. 
While the model performs well across a broad range of scenarios due to the generalization capacity of its pre-trained CLIP encoder, segmentation accuracy may degrade when the provided text or visual prompts represent concepts that are significantly outside of its training distribution or involve highly specialized contexts, which we believe is the underlying reason for its subpar segmentation performance on debris.

Given these challenges, we propose fine-tuning a pre-trained CLIPSeg model to perform debris segmentation from post-hurricane aerial imagery. 
Utilizing a foundational vision model allows us to reach the desired segmentation performance with a relatively small fine-tuning dataset.
Specifically, we curate a dataset of approximately 1,200 images from hurricanes Ian, Ike, and Ida.
The dataset contains a balanced mix of debris-present and debris-free examples to reduce class imbalance.
All images captured in the Hurricane Ida region are withheld during training to evaluate domain generalization.
Each image in this dataset is labeled by multiple annotators; we aggregate their labels to generate consensus annotations to reduce human biases and errors.
We also employ techniques such as visual prompt engineering to highlight salient debris visual features, further boosting training efficiency.
Through quantitative and qualitative evaluations, we demonstrate that fine-tuning CLIPSeg using our curated debris dataset leads to significant increases in debris segmentation performance.
Our experiments indicate that our model is robust to changes in environment and imaging factors, making it the first segmentation model applicable to multiple disaster regions.

The ultimate goal of our work is to enable near-real-time debris detection that supports rapid disaster response and informs long-term resilience planning. By leveraging publicly available aerial imagery, such as NOAA Emergency Response Imagery, our framework aims to offer a broadly applicable and reproducible solution across diverse disaster contexts. Beyond immediate operational benefits—like prioritizing cleanup and assessing disaster impacts—this approach lays the groundwork for future research in predictive modeling and multi-scale debris analysis. By committing to open-sourcing the implementation and annotated dataset \footnote{Source code and dataset will be made publicly available upon acceptance of the peer-reviewed article.}, we seek to foster collaboration and continuous innovation, empowering researchers and practitioners to enhance disaster preparedness and recovery efforts worldwide.
We summarize our key contributions as follows:
\begin{itemize}  
    \item A high-quality dataset of 1,200 labeled debris images sampled from Hurricanes Ian, Ida, and Ike is provided.
    \item Human annotation biases are mitigated by aggregating labels from multiple annotators into a consensus segmentation, ensuring more reliable ground truth for model fine-tuning.
    \item Foundational vision models are fine-tuned to achieve effective debris segmentation from aerial imagery.
    \item Experimental results demonstrate robust performance of the segmentation model on disaster regions unseen during training and resilience to varying environmental and imaging conditions.
    \item This model represents the first debris segmentation solution applicable across multiple disaster regions.
    \item Both the model and dataset are open-sourced to empower disaster recovery efforts.
\end{itemize}

\begin{figure*}
	\centering
	\includegraphics[width=1.0\linewidth]{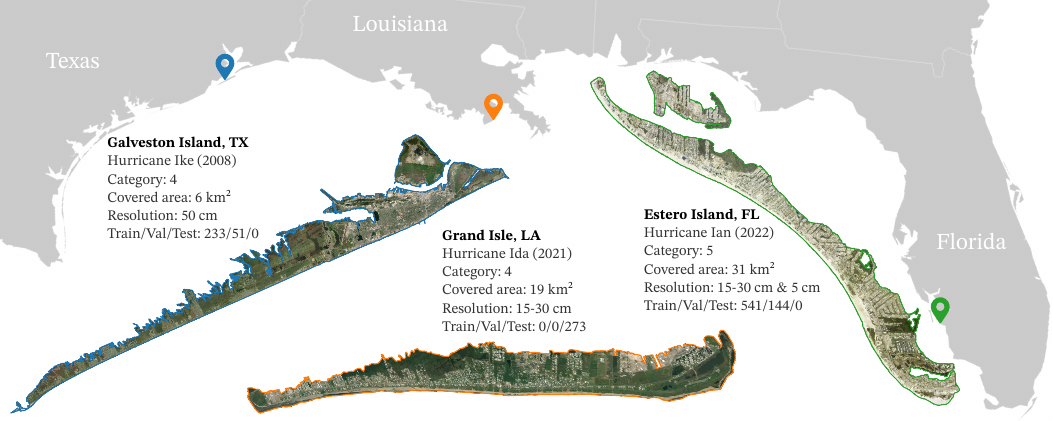}
	\caption{\textbf{Three hurricane-affected regions used in this research.}
    Key details for each storm and location are provided, alongside visualizations highlighting the most significantly impacted areas and the number of labeled images in the training, validation, and test sets.
    The diversity of conditions across regions is crucial for training and evaluating segmentation models.}
	\label{fig_vis_case_studies}
\end{figure*}
\section{Related Work} \label{sec:related_work}

Debris detection has been widely studied in environmental monitoring and remote sensing \citep{cheng2024post, cheng2024framework}, yet its application in civil engineering—particularly for assessing impacts on the built environment—remains relatively under-explored. Debris detection can be divided into several sub-categories depending on the context and the surrounding environment, each requiring tailored methods to capture its unique characteristics. For example, in the context of \textit{marine debris}, the task is to identify persistent solid materials that are discarded or abandoned in aquatic settings \citep{hu2021remote}. Existing studies typically classify marine debris pixels based on spectral signatures—the variation in reflectance or emittance of a material across wavelengths. Leveraging multi-spectral or hyper-spectral imaging instruments, machine learning algorithms with relatively low model capacity, such as Naïve Bayes \citep{biermann2020finding}, Random Forest \citep{jamali2021cloudbased}, and Support Vector Machines (SVM) \citep{jamali2021cloudbased}, have been effectively employed to distinguish debris (e.g., macro-plastics) from the surrounding marine background.

\textit{Vegetative debris}, on the other hand, consists of whole trees, tree stumps, tree branches, tree trunks, and other leafy materials \citep{fema_vegetative}.
The study of vegetative debris often focuses on its impact on infrastructure, such as roadways, where the spectral contrast between vegetation and background materials like asphalt and concrete can aid in detection. 
Therefore, SVMs can be used to classify vegetative debris using the changes in spectral contrast between pre-hurricane and post-hurricane multi-spectral images \citep{karaer2021posthurricanea, kim2024assessing}.

In this study, we specifically target identifying hurricane-induced debris on land, focusing primarily on materials resulting from damage to buildings and other structures, excluding vegetation.
Many existing methods in this domain also rely on single-pixel classification techniques, using differences in spectral signatures to detect debris from other object categories. 
For example, \citet{hauptman2024hurricane} used a maximum likelihood classification trained on 30 samples (all of which collected in a single region after Hurricane Ian) to classify debris pixels.
However, unlike marine or vegetative debris, non-vegetative debris poses significant detection challenges due to its fragmented and heterogeneous nature, coupled with the lack of consistent visual patterns.
In urban environments, where the background predominantly consists of non-vegetative elements, the spectral contrast is insufficient to effectively differentiate between damaged structures, such as debris, and intact structures.
As such, single-pixel classification algorithms that rely on spectral signatures lead to subpar performance \citep{hauptman2024hurricane}.
\citet{jiangAutomaticUrbanDebris2016} used textural filters in addition to single-pixel spectral classification to delineate debris zones.
These filters are \textit{hand-crafted}; they are effective in controlled settings (i.e. within the training distribution) but are sensitive to variations in imaging and environmental conditions, raising questions about their adaptability to other disaster regions.
To the best of our knowledge, there is no debris segmentation model that is generalizable to different natural hazard events and locations.

In contrast to traditional methods that rely on single-pixel spectral classification or handcrafted textural filters, leveraging high-level vision features—such as the recognition of textures, objects, anomalies, and contextual scenes—can significantly enhance the detection of post-hurricane debris via remote sensing.
High-level vision features necessitate models with substantially more parameters, which in turn require substantial amounts of training data to achieve optimal performance. 
We circumvent the issue of data scarcity by leveraging the pre-trained knowledge and generalizability of foundational models, and fine-tune using a relatively small, high-quality debris dataset.
Additionally, we aim to develop an algorithm that relies solely on aerial RGB images, eliminating the need for ancillary datasets demanding additional hardware, such as LiDAR or multi-spectral imaging instruments. While this approach increases the problem complexity, it enhances the algorithm's transferability and facilitates deployment across diverse regions, making the solution more adaptable to various operational environments. 

\section{Debris segmentation dataset} \label{subsec:debris_dataset}

In this section, we describe the process of building a dataset to fine-tune the foundation image segmentation model, CLIPSeg. Our approach begins with data collection from publicly available aerial images captured in hurricane-impacted regions, followed by debris classification and segmentation annotation. We sample crops from the RGB images at a constant ground resolution of \SI{50}{\meter} $\times$ \SI{50}{\meter}. This fixed dimension is chosen for two reasons: it guarantees compatibility with the optimal resolution range of CLIPSeg's ViT-B/16 encoder ($200\times200$ to $350\times350$ pixels) \citep{luddecke2022image}, and (2) it produces images that are ideally sized for manual annotation—large enough to capture fine debris details without requiring annotators to zoom in excessively. The procedures outlined here are region-agnostic and applicable across diverse geographical contexts affected by hurricanes.

\subsection{Data collection} \label{subsec:data-collection}

A key goal of this work is to develop a generalizable debris detection framework and release both the model and dataset as open-source resources for broader community use.
Our public dataset is available on the NHERI DesignSafe Cyberinfrastructure \citep{DesignSafe_2017}.
With generalizability in mind, we evaluated various publicly accessible imagery sources to meet three principal criteria: 
(1) the data must be promptly and consistently collected following a major hurricane event, (2) the imagery must have sufficient quality for reliable debris detection, and (3) the dataset must be freely and openly available to both researchers and practitioners. 

\subsubsection{Post-hurricane aerial imagery} \label{subsubsec:aerial-imagery}

After assessing multiple candidates, we selected the \emph{NOAA Emergency Response Imagery} dataset \citep{noaa_emergency_response_imagery}. 
This program has been operational since 2003 and continuously collects high-quality aerial photographs of regions impacted by major hurricanes (see \cref{subsubsec:case-studies} for details on data type, quality, and instrument). 
These images are typically made available to the public within a relatively short time—often within days—after the hurricane has made landfall. 
As a result, emergency responders, local governments, and researchers can swiftly access and use these images for immediate post-disaster assessments.


While the NOAA Emergency Response Imagery dataset serves as our primary imagery source, we emphasize that the proposed debris segmentation model and methodology can be applied to aerial imagery from other sources, locations, and resolutions as well, which we discuss in \cref{subsec:generalizability}. 

\subsubsection{Case studies} \label{subsubsec:case-studies}

To train and evaluate our proposed approach, we consider three major hurricane events: \emph{Ian}, \emph{Ida}, and \emph{Ike}. 
These were selected because they span different years, regions, and imaging instruments, thereby testing the capacity of our framework to handle variations in spatial resolution and diverse environmental settings. 
Specifically, we  train the model using images from Hurricanes Ian and Ike and then evaluate on the test set, which exclusively comprises of images from Hurricane Ida. 
This experimental design showcases the generalizability of our model to unseen areas and distinct event characteristics.

\cref{fig_vis_case_studies} illustrates the main locations under study, highlights regions with the most significant debris impact for each hurricane, and displays the number of training, validation, and testing sample images for each event. 
Below is an overview of the key study areas:
\begin{itemize}
    \item \textbf{Hurricane Ian (2022):} 
    Estero Island and Fort Myers in Florida, covering roughly 31~km$^2$. 
    Resolutions here range from 15--30~cm for aerial imagery from NOAA and can reach up to 5 cm for drone-based datasets collected with a longer delay \citep{ocm_partners_2025}. These areas were severely impacted by storm surge and high winds, causing extensive debris.
    
    \item \textbf{Hurricane Ida (2021):} 
    Grand Isle, Port Fourchon, and Golden Meadow in Louisiana. 
    This region encompasses barrier islands and coastal towns severely affected by storm surge and high winds. 
    The total coverage is approximately 19~km$^2$, with resolutions of around 15--30~cm. 
    
    \item \textbf{Hurricane Ike (2008):} 
    Galveston Island in Texas, with approximately 6~km$^2$ coverage. 
    The imagery, captured at roughly 50 cm resolution using older imaging instruments, broadens the model’s exposure to varied conditions and fosters more robust learning.
\end{itemize}

While only certain high-debris regions are explicitly visualized in \cref{fig_vis_case_studies}, our dataset also includes broader surrounding areas and less impacted inland locations for completeness. 
Overall, these three hurricanes provide diverse testbeds for assessing resolution, sensor variety, and debris conditions, thereby confirming the adaptability of our framework.

\subsection{Debris classification} \label{subsec:debris_classification}

\begin{figure}[h]
	\centering
	\includegraphics[width=1.0\linewidth]{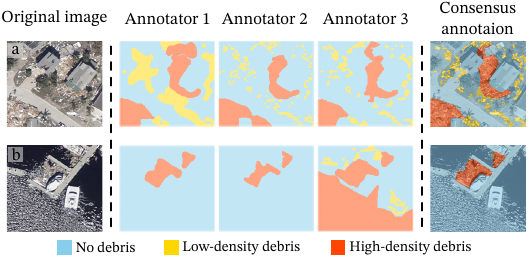}
	\caption{ \textbf{Debris segmentation is challenging – even humans make mistakes.}
    We aggregate inputs from three annotators to obtain a consensus annotation, which is less error-prone.
    Our aggregation strategy is effective at resolving minor disagreements between annotators (row (a)) and rectifying apparent human errors (Annotator 3 in row (b)).
    We use consensus annotations as ground truth labels for fine-tuning.}
	\label{fig_label_agg}
\end{figure}

A training set composed exclusively of debris-laden areas can bias the model toward predicting debris everywhere. To mitigate such bias, the dataset includes crops from non-debris regions—such as coastal areas, forests, built-up zones unaffected by hurricanes, and waterways—to provide negative examples and help balance the model’s perspective.
A multi-annotator manual classification is performed to filter out any image that contains debris.
We call images without any debris (i.e., debris-free) \textit{negative} samples, while images containing debris are called \textit{positive} samples.
Some negative samples contain objects that are close to debris in terms of visual appearance, for example, water reflections, rooftops, rocks, and cemeteries.
The inclusion of positive and negative examples facilitates \textit{contrastive learning} \citep{khosla2020supervised}, or the learning of dissimilarity between debris and other objects.
Query images that are classified as debris-positive are then sent to the segmentation labeling process in \cref{subsec:labeling-procedure}.

\subsection{Debris segmentation annotation} \label{subsec:labeling-procedure}

Consider an arbitrary RGB image $Q \in \mathbb{R}^{H \times W \times 3}$, a human annotator produces a multi-class dense segmentation annotation $S_a \in \{ 0, 1, 2\}^{H \times W}$. Each pixel of $S_a$ is labeled as 0 (no debris), 1 (debris at low-density), or 2 (debris at high-density). For this study, we define \textit{low-density debris} as scattered fragments that exhibit noticeable gaps between pieces, whereas \textit{high-density debris} refers to heavily aggregated piles that typically accumulate considerable mass or height. While the boundary between these two categories is inherently subjective, this distinction is practical for guiding response and cleanup efforts. Low-density debris appears more dispersed, posing fewer immediate threats to the built environment, whereas high-density debris can obstruct critical roadways or facilities and often demands unique strategies for volume estimation. Moreover, separating debris into these two categories leverages CLIP’s capacity to discern distinct visual patterns, improving detection accuracy. Nevertheless, in applications where debris density distinctions are unnecessary, both density classes may be merged into a single debris label for simpler downstream applications.

Relying on segmentation annotation from a single annotator is prone to introduce biases to the dataset, such as tendencies to either under- or over-label, or failure to distinguish debris from visually similar objects (e.g., water reflections in \cref{fig_label_agg}). 
To mitigate human biases in the annotation process, we assign a pool of $N$ annotators $\mathcal{A} = \{a_1, ..., a_N \}$ to each query image $Q$, resulting in $N$ annotations $\{ S_{a_1}, ..., S_{a_{N}} \}$. Let $S_N \in \mathbb{Z}^{H \times W \times N}$ be a 3D tensor of stacked individual annotations. 
We aggregate annotations by acquiring the average label from the pool of annotators.
This is achieved by taking a per-pixel average:

\begin{equation}
	\bar{S}(:, :) =  \lceil \frac{1}{N} \sum_{a=1}^{N} S_N(:, :, a) \rceil,
	\label{eq_label_agg}
\end{equation}

\noindent where the ceiling operator $\lceil x \rceil$ rounds to the smallest integer greater than $x$. We call the aggregated annotation $\bar{S} \in \mathbb{Z}^{H \times W}$ the \textit{consensus annotation}, which we use as ground truth to fine-tune CLIPSeg.

\cref{fig_label_agg} shows two representative examples of label aggregation in \cref{eq_label_agg} correcting biases from individual annotators. In \cref{fig_label_agg}(a), there is a minor disagreement between annotators: annotator 1 tends to over-annotate, whereas annotator 3 tends to under-annotate. The consensus annotation averages over the two biased tendencies. In \cref{fig_label_agg}(b), annotator 3 makes an apparent error, falsely labeling water reflections as low-density debris. Our aggregation strategy corrects such individual errors, resulting in a consensus annotation that is generally more reliable and accurate than any single annotator’s contribution.

\section{Debris Segmentation Model}\label{sec:methodology}

\begin{figure}[ht] 
	\centering
	\includegraphics[width=1.0\linewidth]{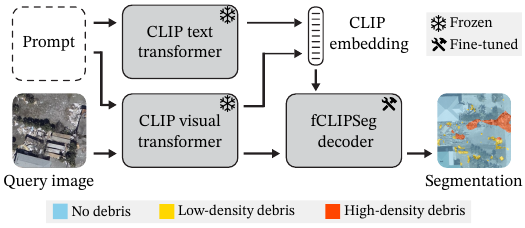}
  \caption{\textbf{Overview of our debris segmentation model, named fCLIPSeg.} Built on the same architecture as the non-fine-tuned CLIPSeg, our model
uses a pair of pre-trained CLIP visual-text encoders, and a decoder fine-tuned on labeled debris images.
    fCLIPSeg produces a segmentation output for each query RGB image. See legend for debris density levels.}
	\label{fig_clipseg}
\end{figure}

Having acquired labeled debris images in \cref{subsec:debris_dataset}, we are now ready to fine-tune the segmentation model CLIPSeg to improve segmentation performance on debris images. 
In this section, we briefly describe how we adopt a standard CLIPSeg model to perform segmentation, followed by our proposed procedures to fine-tune CLIPSeg.

\subsection{Image segmentation using CLIPSeg} \label{subsec:clipseg}

Given an RGB query image $Q \in \mathbb{R}^{H \times W \times 3}$, the model produces a segmentation $S \in \mathbb{Z}^{H \times W}$, where $H$ and $W$ correspond to the height and width of the image, respectively.
We expect the segmentation output to be a \textit{dense} segmentation, where every pixel is associated with a label $L \in \{ 0, 1, 2 \}$ corresponding to text prompts: `no debris,'' `debris at low-density,'' and `debris at high-density,'' respectively (see \cref{subsec:labeling-procedure} for definitions). Collectively, these are referred to as candidate text prompts.

We build our debris segmentation model following the network architecture of CLIPSeg \citep{luddecke2022image}, as seen in \cref{fig_clipseg}.
CLIPSeg uses a visual transformer-based CLIP  encoder (specifically, ViT-B/16 with a patch size $P = 16$) \citep{radford2021learninga} to process image prompts and queries.
An additional CLIP transformer-based text encoder processes text prompts.
CLIPSeg utilizes text and visual CLIP encoders as frozen feature extractors, mapping text and visual prompts to their CLIP embeddings $\bv{e} \in \mathbb{R}^{512}$.

After encoding, a transformer-based decoder follows to accomplish segmentation.
As the query image $Q$ traverses the CLIP visual transformer, activations from selected encoder layers $\{3, 7, 9\}$ are extracted and projected onto a $D$-dimensional token embedding space in the decoder, with $D=64$ in our implementation.
These projected activations, including the CLS token, are combined with the decoder's internal activations at the input of each transformer block.
Ultimately, a linear projection is applied to the decoder's final-layer tokens, converting their shape from $\mathbb{R}^{(1 + \frac{H}{P} \times \frac{W}{P}) \times D}$ to a single-channel segmentation map $S_{l} \in \mathbb{R}^{H \times W}$ corresponding to density level $l$. 

At this stage, the pixel values in $S_{l}$ can be interpreted as the degree to which the pixel matches its input prompt (either text or image, or both). 
To obtain multi-class dense segmentation $S$, we generate $S_{l}$ for all debris densities $l \in L$.
At each pixel location $(x,y)$, label prediction is given by the label with the highest logit value among the outputs $S_l(x,y)$:

\begin{equation}
	S(x, y) = \arg\max_l S_l(x, y),
	\label{eq_multi_class}
\end{equation}

We apply the above procedure to segment debris in aerial images. However, the model struggles with debris compared to other common objects (e.g., cars and buildings), likely because debris is either missing or under-represented in the original training dataset used for the CLIP encoder and CLIPSeg decoder.
In order to adopt CLIPSeg to accurately perform segmentation on a new, complex class of objects (in this case, debris) that is likely unseen during training, we propose to fine-tune CLIPSeg on a debris dataset, which we describe in \cref{subsec:debris_dataset}.

\subsection{Visual prompt engineering} \label{subsubsec:vis_eng}

\begin{figure}
	\centering
	\includegraphics[width=1.0\linewidth]{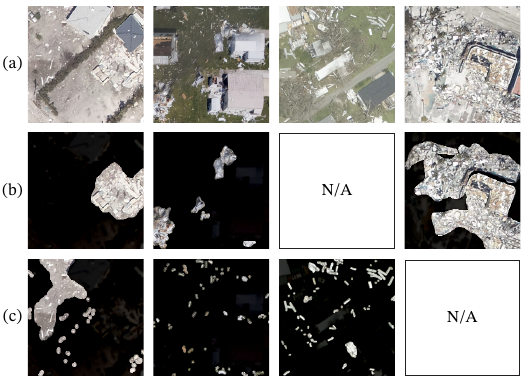}
	\caption{\textbf{Visual prompt engineering for debris images.} We highlight salient debris objects by blurring and darkening background to improve training efficiency. Given (a) RGB images, we generate engineered visual prompts for (b) high-density and (c) low-density debris, respectively (except when the specified density is absent in the RGB images, indicated as N/A). }
	\label{fig_vis_prompt_eng}
\end{figure}

Training a segmentation model to handle complex visual queries such as debris is challenging, especially with limited data.
To improve training efficiency, we employ 
\textit{visual prompt engineering} \citep{shtedritski2023does}, a technique of modifying the RGB images (called visual prompts) by highlighting salient targets (in this case, debris).
It has been demonstrated that visual prompt engineering improves the alignment (measured cosine similarities) between visual and text prompts \citep{luddecke2022image}, which in turn leads to increased segmentation performance.

We highlight salient pixels in an RGB image by artificially degrading the unrelated objects (called background) in the image.
This is achieved by (a) decreasing the background brightness, and (b) applying a Gaussian blur to the background.
We use the debris density labels to define foreground and background pixels.
To obtain the engineered visual prompt for low-density debris, for example,  we darken and blur the background, which are the areas not labeled as low-density debris. 
We repeat the same procedure to obtain the engineered visual prompt for high-density debris.
Visual prompt engineering is not applied to debris-free images.
See \cref{fig_vis_prompt_eng} for illustration.

We call the set of all engineered visual prompts for low and high-density $\mathcal{P}_1$ and $\mathcal{P}_2$, respectively.
The set of visual prompts for no debris $\mathcal{P}_0$ are non-engineered RGB images classified as negative (debris-free) in \cref{subsec:debris_classification}. These engineered datasets are subsequently used for fine-tuning the model, ensuring that both visual and textual prompts are optimally aligned for accurate debris segmentation.

\subsection{Fine-tuning CLIPSeg} \label{subsubsec:fine-tune}

In this section, we detail the fine-tuning process for the transformer decoder illustrated in \cref{fig_clipseg}. 
We initialize fine-tuning with the official CLIPSeg model checkpoint while keeping the CLIP vision and text encoders frozen throughout training. 
Fine-tuning requires a dataset of labeled debris images, where each RGB query image \(Q\) is paired with engineered visual prompts \(\mathcal{P}\) and its corresponding consensus annotation \(\bar{S}\), which we use as ground truth. 

For an arbitrary query image $Q$, we randomly sample a density level $l \in L$ with the corresponding text prompt from "no debris", "debris at low-density", and "debris at high-density".
For the given text prompt, we randomly sample an engineered visual prompt from $\mathcal{P}_l$ (see \cref{subsubsec:vis_eng} for definition and \cref{fig_vis_prompt_eng} for examples).
Ground truth segmentation is obtained by parsing the ground truth segmentation $\bar{S}$ to obtain a binary annotation corresponding to the density level $l$.
After CLIP encoding of the text and visual prompt, we obtain their CLIP embeddings, $\bv{e}_\text{t}$ and $\bv{e}_\text{v}$, respectively.
Because CLIP embeds both images and text into a common latent space, interpolation between their embeddings is feasible:

\begin{equation}
	\bv{e}_\text{c} = \alpha \bv{e}_\text{t} + (1 - \alpha ) \bv{e}_\text{v},
	\label{eq_text_img_interp}
\end{equation}
where $\alpha$ is uniformly sampled between $[0, 1]$.
We use the randomized interpolated conditional embedding $\bv{e}_\text{c}$ as a data augmentation strategy during training.
We use binary cross-entropy as the only loss function, calculated between CLIPSeg output for the corresponding debris density level and the ground truth binary segmentation.
We call the resulting fine-tuned CLIPSeg model \textit{fCLIPSeg} (\textit{f} stands for fine-tuned).
\section{Results} \label{sec:results}
In this section, we present the experimental procedures, quantitative results, and qualitative assessments of our method.
First, \cref{subsec:training-testing} outlines the experimental setup used for training and testing our model. 
Then, We continue with a quantitative evaluation in \cref{subsec:quantitative-evaluation}, where we analyze how well our fine-tuned model performs under various metrics. 
Finally, \cref{subsec:qualitative-evaluation} offers visual examples to highlight the strengths and potential limitations of our approach. 

\subsection{Experimental setup} \label{subsec:training-testing}

We used PyTorch Lightning for training and evaluation.
We fine-tuned CLIPSeg on our debris dataset using a batch size of $64$ for $2000$ epochs, resumed from checkpoint \textit{rd64-uni-refined} released by \citet{luddecke2022image}.
We optimized our model with AdamW \citep{loshchilov2017decoupled}, starting with a learning rate of $0.001$, gradually decaying to $0.0001$ using a cosine scheduler without warmup.
Automatic mixed precision was employed throughout training.
Training took $5$ hours on a single Nvidia A40 GPU with $48$ GB of VRAM.
We used Dice score (debris-classes only) computed on the validation set as the model checkpoint selection criterion.

Since we needed to iterate through all density levels, we ran inference on each query image sequentially.
The effective batch size was $3$ during inference, determined by the number of density levels.
Inference throughput on a single Nvidia RTX 3090 GPU was $9.6$ images per second, although there was room for additional engineering to increase inference-time batch size for further speedup.
The lightweight nature of fCLIPSeg made it efficient to deploy over large geographical regions even with fairly moderate compute hardware, especially with the small batch size at inference time, where peak VRAM consumption was only $1.2$ GB.
For instance, applying fCLIPSeg to the entire Estero Island, TX (covering approximately 10 km\textsuperscript{2} with 4,016 images) required 2 minutes and 19 seconds on hardware consisting of an Intel\textsuperscript{\textregistered} Core\textsuperscript{\texttrademark} i9-13900K CPU and an Nvidia RTX 4090 GPU (24GB VRAM).

\subsection{Quantitative evaluation} \label{subsec:quantitative-evaluation}

\begin{figure*}
	\centering
	\includegraphics[width=1.0\linewidth]{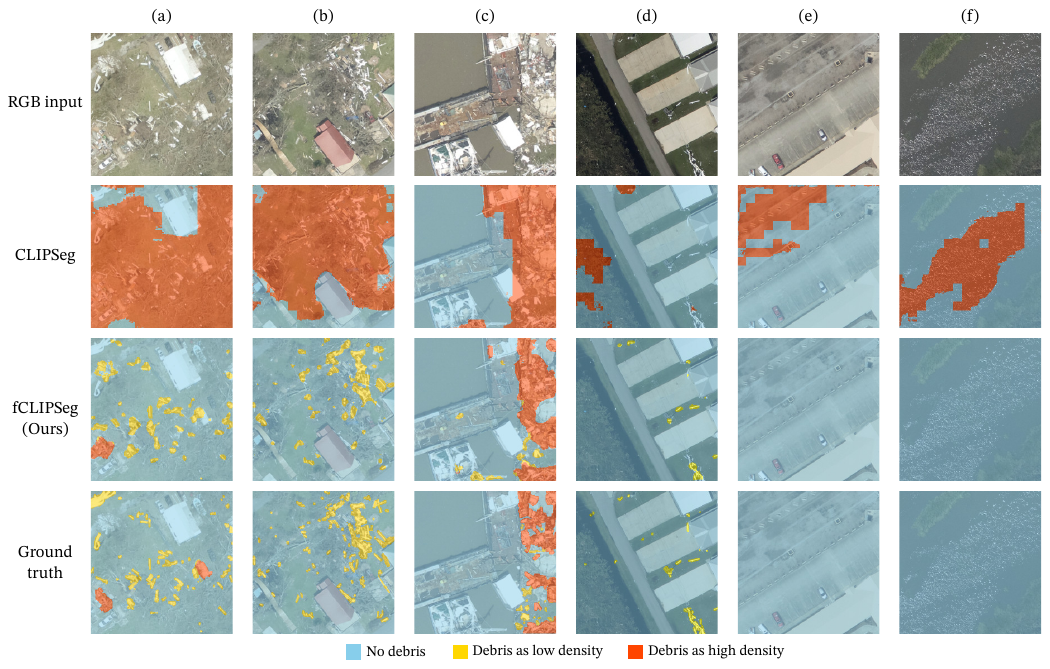}
	\caption{
	\textbf{Visual comparison of debris segmentation models.}
    This figure displays six examples from the test set (Hurricane Ida, 2021): panels (a)–(d) are from the debris-positive subset, while (e)–(f) are from the debris-free subset. fCLIPSeg accurately segments debris without erroneously classifying other objects as debris, unlike its non-fine-tuned variant, CLIPSeg.}
	\label{fig_vis_compare}
\end{figure*}

\begin{table}[ht]
\centering
\caption{\textbf{Segmentation performance across the test set}. The test set comprises 273 images collected from Grand Isle, LA, after Hurricane Ida (2021).
An upper arrow ($\uparrow$) indicates higher values for this metric correspond to better performance.
}
\footnotesize
\label{tab:model_comparison}
\begin{tabular}{@{}llcc@{}}
    &&&fCLIPSeg\\
    Subset & Metric & CLIPSeg & (Ours) \\
	\toprule
	\multirow{6}{*}{Debris-positive} & Dice $\uparrow$ & 0.30 & \textbf{0.70} \\
	& IoU $\uparrow$ & 0.27 & \textbf{0.65} \\
	& Recall [low-density] $\uparrow$ & 0.00 & \textbf{0.33} \\
	& Recall [high-density] $\uparrow$ & \textbf{0.92} & 0.82 \\
	& Precision [low-density] $\uparrow$ & \textbf{0.86} & 0.60 \\
	& Precision [high-density] $\uparrow$ & 0.10 & \textbf{0.87} \\
	\midrule
	\multirow{3}{*}{Debris-free} & Dice $\uparrow$ & 0.74 & \textbf{0.99} \\
	& IoU $\uparrow$ & 0.74 & \textbf{0.99} \\
	& Recall [no debris] $\uparrow$ & 0.93 & \textbf{1.00} \\
	\bottomrule
\end{tabular}
\end{table}

To evaluate model performance, we compare segmentation outputs against consensus annotations obtained from \cref{subsec:labeling-procedure} as ground truth.
We conduct evaluation on the test set, consisting of $273$ images acquired from Grand Isle, Port Fourchon, and Golden Meadow in Louisiana after Hurricane Ida, where both regions are unseen during training. We further split the test set into two subsets, following the binary manual classification result described in \cref{subsec:debris_classification} : (a) debris-positive subset, with $125$ samples, and (b) debris-free (negative) subset, with $148$ images.

Evaluation results are shown in \cref{tab:model_comparison}.
\textit{Precision} quantifies the proportion of correctly identified positive pixels (true positives) among all pixels predicted as positive, reflecting the accuracy of the positive predictions. 
\textit{Recall}, on the other hand, measures the proportion of true positive pixels identified by the model out of all actual positive pixels in the ground truth, indicating the model's ability to capture all relevant pixels.
\textit{Dice Similarity Coefficient} \citep{bertels2019optimizing} is the harmonic mean of precision and recall, measuring the balance between false positives and false negatives; 
\textit{Intersection over Union (IoU)} quantifies the overlap between the predicted segmentation and the ground truth by calculating the ratio in area of their intersection to their union.

We compare our debris segmentation model, fCLIPSeg, against CLIPSeg \citep{luddecke2022image} without fine-tuning.
The two models are otherwise identical.
For the debris-positive subset, Dice and IoU scores of CLIPSeg suggest that prior to fine-tuning, CLIPSeg has some rudimentary capacity for debris segmentation.
Although recall and precision scores suggest that CLIPSeg is highly biased; 
very low recall for low-density debris reveals that CLIPSeg is not sensitive to low-density debris, unable to capture all relevant pixels, while very low precision for high-density debris reflects on the poor accuracy of positive prediction for this label class.

Fine-tuning of CLIPSeg significantly improves the debris segmentation performance, as seen in the improved Dice and IoU scores in \cref{tab:model_comparison}.
Improved recall for low-density suggests that fCLIPSeg can detect more relevant pixels in this class, while improved precision for high-density shows that fCLIPSeg is far more accurate in identifying high-density debris.
Fine-tuning improves the overall accuracy and reduces bias in debris segmentation. For the debris-free subset, Dice and IoU scores for fCLIPSeg approach $1.0$, suggesting that the model rarely misclassifies non-debris objects as debris, while CLIPSeg frequently makes errors. 
Overall, these quantitative metrics lend confidence to the use of fCLIPSeg and highlight the value of fine-tuning, including reduced bias, improved accuracy, and near-perfect performance for debris-free images. 

\subsection{Qualitative evaluation} \label{subsec:qualitative-evaluation}

\cref{fig_vis_compare} shows a visual comparison between CLIPSeg and our model, fCLIPSeg, on a few example images from the test set.
Images (a) -- (d) are sampled from the debris-positive subset, while images (e) -- (f) are sampled from the debris-free subset.
Outputs from CLIPSeg (prior to fine-tuning) suggest that the model can detect debris in some instances but suffers from very poor accuracy.
For example, CLIPSeg falsely labels vegetation in (a), (b), and (d), parking lot markings in (e), and water reflections in (f) as debris. Additionally, high-density (orange) segmentations in (a) and (b) noticeably lack granularity, bleeding into adjacent debris-free pixels.
CLIPSeg's tendency to mislabel other objects as high-density debris (orange) aligns with its low precision score in \cref{tab:model_comparison}.
Conversely, CLIPSeg exhibits the opposite tendency for low-density debris, failing to detect relevant pixels in (d) and falsely categorizing low-density pixels as high-density debris in (a) and (b). This behavior is also reflected in its poor recall score in \cref{tab:model_comparison}.

Across all examples in \cref{fig_vis_compare}, fCLIPSeg provides satisfactory debris segmentation performance.
Low-density debris tend to appear small and fragmented in aerial images, which makes them difficult to segment.
Nonetheless, fCLIPSeg correctly segments almost all low-density debris in (a) and (b), without segmenting the surrounding debris-free pixels.
\cref{fig_vis_compare}(c) shows that fCLIPSeg performs well in heavily damaged areas, correctly labeling high-density debris pixels while excluding regions without debris.
\cref{fig_vis_compare} also reveals that fCLIPSeg is not over-fitted on sparse image features, such as image brightness (most debris objects are white) and shape (many debris fragments are rectangular in shape), but instead relies on high-dimensional CLIP feature space to reliably differentiate debris from other visually similar objects.
This is evident in \cref{fig_vis_compare}(c), where fCLIPSeg labels repeated, well-structured objects correctly, and in (f), where fCLIPSeg labels fragmented water reflections correctly.
fCLIPSeg's high accuracy in identifying non-debris objects is reflected in its high recall score under the debris-free subset in \cref{tab:model_comparison}.

\cref{fig_vis_compare} also reveals that fCLIPSeg can make minor errors.
For example, some small fragments of low-density debris are missing in the fCLIPSeg output (e.g., \cref{fig_vis_compare}(d)).
However, we anticipate that false positives on tiny debris fragments only have small downstream impacts, for example, the estimation of debris volume.
Additionally, there are scenarios where fCLIPSeg correctly assigns the density levels to debris targets that are inconsistent with the ground truth labels (e.g. \cref{fig_vis_compare}(a)).

\begin{figure}[t]
	\centering
	\includegraphics[width=1.0\linewidth]{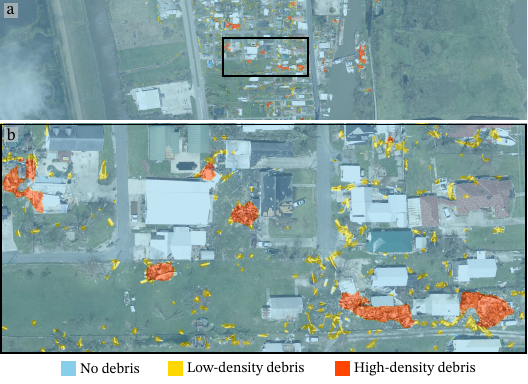}
	\caption{\textbf{Demonstration model deployment over large-scale regions.} fCLIPSeg produces accurate, artifact-free outputs for a region affected by Hurricane Ida in our test set. Panel (b) provides a zoomed-in view of the area enclosed by the black box in (a).}
	\label{fig_regional_result}
\end{figure}

At a regional scale, \cref{fig_regional_result} further demonstrates fCLIPSeg’s capability to accurately identify debris across large swaths of the test area, which is entirely unseen during training. To deploy fCLIPSeg efficiently over large regions, we crop the aerial imagery into $ 50 \times 50$ m patches and compute segmentation for each crop, before merging the individual outputs to form comprehensive regional segmentation maps, such as \cref{fig_regional_result}. This approach enables rapid deployment over large regions (discussed in \cref{subsec:training-testing}) without apparent discontinuities or distortions at patch boundaries.  Panels (a) and (b) in \cref{fig_regional_result} show two different zoom levels in the same region of Louisiana affected by Hurricane Ida, revealing detailed segmentations of both low-density and high-density debris while leaving unaffected areas unmarked. Importantly, the model does not mistakenly label non-impacted zones, suggesting that it can generalize well over broader geographic extents. Overall, this ability to effectively capture debris distributions at multiple spatial scales underscores the practicality of fCLIPSeg for real-world disaster response and recovery operations. 

Lastly, \cref{fig_vis_compare} and \cref{fig_regional_result} only include images sampled from the test set.
See \cref{subsec:generalizability} for discussions on fCLIPSeg's ability to generalize across a variety of conditions.

\section{Discussion and Conclusion} \label{sec:discussion}
This section synthesizes our findings on fCLIPSeg's performance in debris segmentation across diverse real-world conditions while acknowledging its limitations. It further outlines promising directions for future research to refine the model and enhance its applicability in disaster response.

\subsection{Generalizability and transferability} \label{subsec:generalizability}

\begin{figure}
	\includegraphics[width=1.0\linewidth]{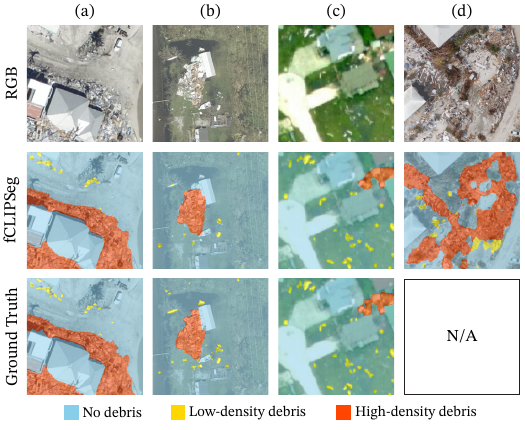}
	\caption{\textbf{Demonstration of segmentation generalizability.}
    fCLIPSeg produces accurate segmentation maps across a variety of conditions: Ian ((a) and (d)), Ida (b), and Ike  (c). Images (a) and (d) are acquired from different imaging instruments.}
	\label{fig_vis_storms}
\end{figure}

Quantitative and qualitative evaluations presented in \cref{subsec:quantitative-evaluation} and \cref{subsec:qualitative-evaluation} were conducted on images collected primarily from Grand Isle, supplemented by additional imagery from Port Fourchon and Golden Meadow, following Hurricane Ida. The entire region in the test set is withheld from being used for training and validation (in particular, hyper-parameter tuning and early stopping) to demonstrate generalizability to unseen conditions.

\cref{fig_vis_storms} further supports fCLIPSeg's generalizability across a variety of conditions, where we show four different samples to illustrate the different conditions that contribute to the diversity in debris image features.
\cref{fig_vis_storms}(a) is captured in Estero Island, FL, after hurricane Ian (Category 5) in 2018.
Strong wind shear and storm surge led to catastrophic damage in the area, and the captured images show significant aggregation of high-density debris.
Due to its proximity to the coastline and the characteristics of the land cover, the image backgrounds, such as roadways, are often covered with sand. This results in limited contrast between the background and the debris in the foreground, making segmentation significantly more challenging compared to other areas.
The image in \cref{fig_vis_storms}(b) is captured in Golden Meadow, LA after Hurricane Ida in 2021, where we observe more vegetation and less severe structural damage. 
\cref{fig_vis_storms}(c) shows an image captured on Galveston Island, TX, after Hurricane Ike in 2008.
At the time, the deployed aircraft were equipped with inferior imaging instruments, and the captured images suffer from poor resolution (at \SI{50}{\centi\meter}), inaccurate color calibration (green color cast).
Despite these challenges, fCLIPSeg still provides an accurate debris segmentation.
In \cref{fig_vis_storms}(d), we present an image captured in the same region as \cref{fig_vis_storms}(a), but with a slight time delay.
While images from \cref{fig_vis_storms}(a) - (c) were captured by aircrafts ($15$ -- \SI{50}{\centi\meter}), \cref{fig_vis_storms}(d) was captured by a drone, and the reduced altitude leads to significantly improved image resolution (\SI{5}{\centi\meter}) and details.
Drone images, such as the one shown in \cref{fig_vis_storms}(d), are not included in our debris dataset. Consequently, no ground truth labels exist for these images, and they were excluded from the model fine-tuning process.
Despite the domain shift in imaging instruments, fCLIPSeg still provides satisfactory segmentation output.

With a few illustrative examples, \cref{fig_vis_storms} showcases a set of factors that might impact debris detection, such as debris density, material type, background color, aircraft altitude, and imaging instrument, among others.
Evaluations reveal that fCLIPSeg is largely not affected by changes across those factors, suggesting a high degree of generalizability.

\subsection{Limitation and challenges}\label{subsec:limitation}
Despite the promising performance of the proposed framework, it is not without shortcomings. 
In some cases, the model is unable to detect every small fragment of debris, especially when debris pieces are very tiny or partially occluded. 
Additionally, fCLIPSeg occasionally misclassifies objects with similar textures (e.g., vegetation or small, regularly shaped man-made objects) as debris.

Some of these misclassifications can be observed in \cref{fig_regional_result}. 
Although such errors are relatively rare, they highlight the potential for further improvements, particularly in refining the model’s ability to handle ambiguous visual cues.

Another challenge stems from the regional and environmental diversity of post-hurricane conditions. 
While the current training dataset covers several hurricanes in the United States, incorporating images from more diverse geographic regions—potentially including different countries with unique construction materials, natural environments, and climatological features—could enhance the model’s generalizability. 
Extending the dataset to capture these variations may help reduce misclassifications and improve the robustness of the model under drastically different scenarios.

\subsection{Future opportunities}\label{subsec:future-works}
Building upon the present framework, several avenues for future work emerge. 
First, although our approach primarily focuses on near-real-time debris detection immediately after a hurricane, an interesting extension would be developing \emph{predictive} models that estimate future debris fields even before a storm makes landfall. 
Such a predictive system could leverage the results of our current model as ground truth to train a lower-resolution, wide-area forecasting model, thereby supporting more proactive emergency planning and risk mitigation.

Second, there is a growing need for three-dimensional (3D) analysis in disaster response. 
By integrating LiDAR data, stereo photogrammetry, or other 3D information, the same debris segmentation pipeline could help produce volumetric or height-based estimates of debris. 
Such a capability would enable a more precise assessment of impacted areas, providing crucial data for emergency management and debris-removal logistics which are often driven by volume of debris.

Lastly, extending our work to generate a global map of post-disaster debris could provide a more equitable solution, especially for regions with limited resources. While our current framework is designed for detailed debris detection using high-resolution aerial imagery in the aftermath of disasters, adapting the approach to operate on lower-resolution satellite imagery would enable rapid, global-scale assessments of disaster-induced debris. This global model would use our high-fidelity outputs as reference points, reducing biases linked to localized training datasets and making post-event debris detection technology more broadly accessible to communities worldwide.

\subsection{Conclusion}\label{subsec:conclusion}
This paper introduced a fine-tuned CLIPSeg model, referred to as \emph{fCLIPSeg}, for robust and generalizable debris segmentation in post-hurricane aerial imagery. 
By incorporating multi-annotator consensus labeling, visual prompt engineering, and contrastive learning, the proposed approach demonstrated substantially improved segmentation performance in both debris-positive and debris-free regions. 
In particular, the fCLIPSeg model effectively adapts to different debris densities, instrument resolutions, and environmental conditions, showing promise for large-scale deployment in diverse disaster response scenarios.

Experimental evaluations indicated that our framework achieves significant gains in precision and recall over a baseline CLIPSeg model, validating the benefits of fine-tuning on a domain-specific dataset. 
Moreover, qualitative assessments revealed that the model handles complex backgrounds, partial occlusions, and visually similar objects with reasonable effectiveness.
At this stage, the model still exhibits occasional errors and misclassifications, and these limitations motivate further research into improving generalizability through additional training data from different global regions and more varied acquisition platforms, such as drones and satellites.

Beyond the immediate post-disaster analysis afforded by the newly proposed model, the results here can also serve as a stepping stone for developing predictive debris analysis tools, transitioning to a 3D context for enhanced volume estimation, and even constructing a global debris map to aid communities and emergency management agencies worldwide. 
We anticipate that making this model, associated codebase, and dataset publicly available will spur broader collaboration and innovation, ultimately enhancing disaster preparedness and recovery efforts.

\begin{acknowledgements}
Kooshan Amini and Jamie E. Padgett were partially supported by NSF Award 2429680 and by the Ken Kennedy Institute’s inaugural Research Clusters Initiative. Additionally, the contributions of Yuhao Liu and Ashok Veeraraghavan were supported by NSF Award 2107313. Any opinions, findings, conclusions, or recommendations expressed in this paper are those of the authors and do not necessarily reflect the views of the sponsors. The authors sincerely thank Narges Saeednejad, Andres Calvo, Jainish Patel, and Benjamin Barria Martinez for their valuable contributions as annotators.
\end{acknowledgements}

\section*{Bibliography}
\bibliography{zotero_ref_yuhao,additional_ref,zotero_ref_kooshan}

\end{document}